\documentclass{bmvc2k}

\usepackage{amsmath,amssymb}
\usepackage{verbatim}
\usepackage{multirow}
\usepackage{color,soul}
\usepackage{mathtools}

\DeclarePairedDelimiter{\ceil}{\lceil}{\rceil}

\title{Weakly-Supervised Video Object Grounding from Text by Loss Weighting and Object Interaction}

\addauthor{Luowei Zhou}{luozhou@umich.edu}{1}{}
\addauthor{Nathan Louis}{natlouis@umich.edu}{2}{}
\addauthor{Jason J. Corso}{jjcorso@umich.edu }{1}{,2}

\addinstitution{
 Robotics Institute\\
 University of Michigan\\
 Ann Arbor, USA
}
\addinstitution{
 Electrical Engineering and Computer Science\\
 University of Michigan\\
 Ann Arbor, USA
}

\runninghead{Zhou et al.}{Weakly-Supervised Video Object Grounding from Text}

\def\eg{\emph{e.g}\bmvaOneDot}

\def\etal{\emph{et al}\bmvaOneDot}
\def\ie{\emph{i.e}\bmvaOneDot}

\begin{document}

\setlength{\abovedisplayskip}{3pt}
\setlength{\belowdisplayskip}{3pt}

\maketitle


\begin{abstract}
We study \textit{weakly-supervised video object grounding}: given a video segment and a corresponding descriptive sentence, the goal is to localize objects that are mentioned from the sentence in the video. During training, no object bounding boxes are available, but the set of possible objects to be grounded is known beforehand.
Existing approaches in the image domain use Multiple Instance Learning (MIL) to ground objects by enforcing matches between visual and semantic features.
A naive extension of this approach to the video domain is to treat the entire segment as a bag of spatial object proposals.
However, an object existing sparsely across multiple frames might not be detected completely since successfully spotting it from one single frame would trigger a satisfactory match. To this end, we propagate the weak supervisory signal from the segment level to frames that likely contain the target object.
For frames that are unlikely to contain the target objects, we use an alternative penalty loss. We also leverage the interactions among objects as a textual guide for the grounding. We evaluate our model on the newly-collected benchmark YouCook2-BoundingBox and show improvements over competitive baselines. 
\end{abstract}

\section{Introduction}\label{sec:intro}

Grounding language in visual regions provides a fine-grained perspective towards visual recognition and has become a prominent research problem in the computer vision and natural language processing communities~\cite{rohrbach2017generating, rohrbach2016grounding, xiao2017weakly, huang2018findingit}. In this paper, we study the problem of \textit{video object grounding}, where a video (segment) and an associated sentence are given and the goal is to localize the objects that are mentioned in the sentence in the video. This task is often formulated as a \textit{visual-semantic alignment} problem~\cite{karpathy2015deep} and has broad applications including retrieval~\cite{karpathy2015deep,karpathy2014deep}, description generation~\cite{yu2013grounded, rohrbach2017generating}, and human-robot interaction~\cite{al2017natural, thomason2017opportunistic}.

Like most fine-grained recognition problems~\cite{ren2017faster, plummer2015flickr30k}, 
grounding can be extremely data intensive, especially in the context of unconstrained video. On the other hand, video-sentence pairs are easier to obtain than object region annotations (\eg, YouTube Automatic Speech Recognition scripts). We focus on the weakly-supervised version of the grounding problem where the only supervision is sentence descriptions; no spatially-aligned object bounding boxes are available for training. Sentence grounding can involve multiple interacting objects, which sets our work apart from the relatively well-studied weakly-supervised object localization problem, where one or more objects are localized independently~\cite{prest2012learning,kwak2015unsupervised}.

Existing work on visual grounding falls into two categories: multiple instance learning~\cite{karpathy2015deep,huang2018findingit} and visual attention~\cite{rohrbach2016grounding}. In either case, the visual-semantic similarity is first measured between the target object/phrase and all the image-level, \ie spatial, object region proposals. Then, either a ranking loss or a reconstruction loss---both of which we refer to here as matching losses---measures the quality of the matching.
A naive extension of the existing approaches to the video domain is to treat the entire video segment as a bag of spatial object proposals. However, this presents two issues.
First, existing methods rely on the assumption that the target object appears in \textit{at least one} of the proposal regions. This assumption is weak when it comes to video, since a query object might appear sparsely across multiple frames\footnote{In YouCook2-BoundingBox, the target object appears in 60.7\% of the total frames, on average.} and might not be detected completely. The \textit{segment-level supervision}, \ie object labels, could be potentially strengthened if applied to individual frames.
Second, a video segment can last up to several minutes. Even with temporal down-sampling, this can bring in tens or hundreds of frames and hence thousands of proposals, which compromise the visual-semantic alignment accuracy.


To address these two issues, we propose a frame-wise loss weighting framework for video grounding. We ground the target objects on a frame-by-frame basis. We face the challenge that the segment-level supervision is not applicable to individual frames where the query object is off-screen, occluded, or just not present in the proposals for that frame. 
Our solution is to first estimate the likelihood that the query object
is present in (a proposal in) each video frame. If the likelihood is high, we judge the matching quality mainly on the matching loss. 
Otherwise, we down-weight the matching loss while bringing in a penalty loss. The lower the confidence, the higher the penalty.
%
With the conditioned frame-wise grounding framework, the proposed model can avoid being flooded with massive proposals even when the sampling rate is high and only make predictions for applicable frames.

We propose two approaches to estimate frame-wise object likelihood (confidence) scores. The first one is conditioned on both visual and textual inputs, namely, the maximum visual-semantic similarity scores in each frame. The second approach is inspired by the fact that the combination of objects can imply their order of appearance in the video. For example, when a sequence of objects ``tomatoes'', ``pan'' and ``plate'' appears in the description, the video scene is likely to include a shot of tomatoes being grilled in the pan at the beginning, and a shot of tomatoes being moved to the plate at the end. In the temporal domain, ``pan'' appears mostly ahead of ``plate'' while ``tomatoes'' intersects with both. We implicitly model the object interaction with self-attention~\cite{vaswani2017attention} and use textual guidance to estimate the frame-wise object likelihood.

For evaluation, due to lack of existing video grounding benchmarks, we have collected annotations over the large-scale instructional video dataset YouCook2, which provides over 15,000 video segment-description pairs. We sample the validation and testing videos at 1fps and draw bounding box for the 67 most frequent objects when they are present in both the video segment and the description. We compare our methods against competitive baselines on video grounding and our proposed methods achieve state-of-the-art performances.

Our contributions are twofold: 1) we propose a novel frame-wise loss weighting framework for the video object grounding problem that outperforms competitive baselines; 2) we provide a benchmark dataset for video grounding.


\section{Related Work}
\noindent\textbf{Grounding in Image/Video.} Supervised grounding or referring has been intensively studied~\cite{plummer2015flickr30k, plummer2017conditional, yu2018mattnet} in the image domain. These methods require dense bounding box annotations for training, which are expensive to obtain.
Recently, an increasing amount of attention has shifted towards the weakly-supervised grounding problem ~\cite{rohrbach2016grounding,xiao2017weakly,karpathy2014deep,karpathy2015deep,huang2018findingit},
where only descriptive phrases, no explicit target grounding locations, are made accessible during training. Karpathy and Fei-Fei~\cite{karpathy2015deep} propose to pair image regions to words in a sentence by computing a visual-semantic similarity score, finding the word that best describes the region. Rohrbach \etal~\cite{rohrbach2016grounding} ground textual phrases in images by reconstructing the original phrase through visual attention. 
Yu and Siskind~\cite{yu2017sentence} ground objects from text in constrained videos.
De-An \etal~\cite{huang2018findingit} extend~\cite{karpathy2015deep} to the video domain and further improve the work by modeling the reference relationships among segments.
In this work, we tackle the problem from a novel aspect as fully exploiting the visual-semantic relations within each segment, \ie frame-wise supervisions and object interactions.

\noindent\textbf{Weakly-supervised Object Localization.}
Weakly-supervised object localization has been explored in both the image \cite{cinbis2014multi,divvala2014learning,deselaers2012weakly,song2014learning,oquab2015object} and the video domain~\cite{prest2012learning,kwak2015unsupervised}. Unlike object grounding from text, object localization typically involves localizing an object class or a video tag in the visual content. Existing works in the image domain naturally pursue a multiple instance learning (MIL) approach to this problem. Positive instances are images where the label is present, and negative instances are given as images with the label absent. In the video domain, the existing methods~\cite{prest2012learning,kwak2015unsupervised} approach this problem by taking advantage of motion information and similarity between frames to generate spatio-temporal tubes. Note that these tubes are much more expensive to obtain compared with spatial proposals, hence we only consider the latter option.


\noindent\textbf{Object Interaction.}
Object interaction was initially proposed to detect fine-grained visual details for action detection, such as the temporal relationships between objects in a scene, to overcome changes in illumination, pose, occlusion, etc.
Some works have modeled object interaction using pairwise or higher-order relationships~\cite{ni2016progressively,lea2016segmental,ma2017attend}.
Ni \etal~\cite{ni2016progressively} consolidate object detections at each step by modeling pair-wise object relationships and hence enforce the temporal object consistency in each additional step. Ma \etal~\cite{ma2017attend} implicitly model the higher-order interactions among object region proposals, using groups and subgroups rather than just pairwise interactions. Inspired by recent work~\cite{xiao2017weakly, cirik2018using}, where the linguistic structure of the input phrase is leveraged to infer the spatial object locations, we propose to model object interaction from a linguistic perspective as a textual guidance for grounding.

\section{Methods}
We start this section by introducing some background knowledge. In Sec.~\ref{sec:video_grounding}, we describe the video object grounding baseline. We then propose our framework in Sec.~\ref{sec:loss_weighting} by extending the segment-level object label supervision to the frame-level. Two novel approaches are proposed in judging under what circumstances the frame-level supervision is applicable.

\subsection{Background}
In this section we provide some background on visual-semantic alignment framework (grounding by ranking) and self attention, which are building blocks of our model. \\

\noindent\textbf{Grounding by Ranking.} We start by describing ranking-based grounding approach from~\cite{karpathy2015deep}. Given a sentence description including $O$ query objects/phrases and a set of $N$ object region proposals from an image, the goal is to target each referred object in the query as one of the object proposals. Queries and visual region proposals are first encoded in a common $d$-dimensional space. Denote the object query feature vectors as $\{q_k\}$, $k=1,2,\ldots,O$ and the region proposal feature vectors as $\{r_i\}$, $i=1,2,\ldots,N$. We pack the feature vectors into matrices $Q=(q_1,\ldots,q_O)$ and $R=(r_1,\ldots,r_N)$. The visual-semantic matching score of the description and the image is formulated as:
\begin{equation}\label{eq:matching_score}
S(Q, R)=\frac{1}{O} \sum_{k=1}^O \max_i a_k^i,
\end{equation}
where $a_k^i = q_k^\top r_i$ measures the similarity between query $q_k$ and proposal $r_i$. Defining negative samples $Q'$ and $R'$ as the query and proposal from texts and images that are not paired with $R$ nor $Q$, the grounding by ranking framework minimizes the following margin loss:
\begin{small}
\begin{equation}
L_{rank} = \sum_{R'\neq R}\sum_{Q'\neq Q} [\max(0,S(Q,R')-S(Q,R)+\Delta)+\max(0,S(Q',R)-S(Q,R)+\Delta)],
\end{equation}
\end{small}
where the first ranking term encourages the correct region proposal matching and the second ranking term encourages the correct sentence matching. $\Delta$ is the ranking margin. During inference, the proposal with the maximal similarity score $a_k^i$ with each object query is selected.

\noindent\textbf{Self Attention.} We now describe the scaled dot-product attention model. Define a set of queries $q_j \in \mathbb{R}^d$, a set of keys $k_t \in \mathbb{R}^d$ and values $v_t \in \mathbb{R}^d$, where $j=1,2,\ldots,O$ is the query index, $t=1,2,\ldots,T$ is the key/value index. Given an arbitrary query $q_k$, scaled dot-product attention computes the output as a weighted sum of values $v_t$, where the weights are determined by the scaled dot-products of query $q_j$ and keys $k_t$, as formulated below:
\begin{equation}\label{eq:sdp}
A(q_j, K, V)=\text{Softmax}(q_j^\top K /\sqrt{d})V^\top,
\end{equation}
where the authors pack $k_t$ and $v_t$ into matrices $K=(k_1,\ldots,k_T)$ and $V=(v_1,\ldots,v_T)$, respectively. \emph{Self-attention}~\cite{vaswani2017attention} is a special case of the scaled dot-product attention where the queries, keys and values are all identical. In our case, they are all object encoding vectors and self-attention encodes the semantic relationships among the objects. We adopt a multi-head version of the self-attention layer~\cite{vaswani2017attention, zhou2018end} for modeling object relationships, which deploys multiple paralleled self-attention layers.

\subsection{Video Object Grounding}\label{sec:video_grounding}

We adapt the Grounding by Ranking framework~\cite{karpathy2015deep} to the video domain, and this adaptation will serve as our baseline. Denote the set of $T$ frames in a video segment as $\{f_t\}$ and the object proposals in frame $t$ as $r_i^t$, $i=1,2,\ldots,N$. As before, define the object queries as $q_k$, 
we compute the similarity between the query object and all the proposals $\{r_i^t\}$ in a segment.
Note that the similarity dot product might grow large in magnitude as $d$ increases~\cite{vaswani2017attention}. Hence, we scale the dot-product by $\frac{1}{\sqrt{d}}$ and restrict $a_k^{t,i}$ to be between 0 and 1 with a Sigmoid function. The similarity function and segment-description matching score are then:
\begin{gather}
a_k^{t,i} = \text{Sigmoid}(q_k^\top r_i^t/\sqrt{d}),\quad S(Q, R)=\frac{1}{O} \sum_{k=1}^O \max_{t,i} a_k^{t,i},
\end{gather}
where matrix $R=(r_1^1, \ldots, r_N^1, r_1^2, \dots, r_N^T)$ indicates the pack of all proposal features.

This ``brute-force'' extension of Grounding by Ranking framework to the video domain presents two issues. First, depending on the video sampling rate, the total number of proposals per segment ($T\times N$) could be extremely large. Hence this solution does not scale well to long frame sequences. 
Second, an object existing sparsely across multiple frames might not be detected completely since successfully spotting it from one single frame would trigger a satisfactory match. We explain next how we propagate this weak supervisory signal from the segment level to frames that likely contain the target object.

\begin{figure}[t]
\includegraphics[width=\textwidth]{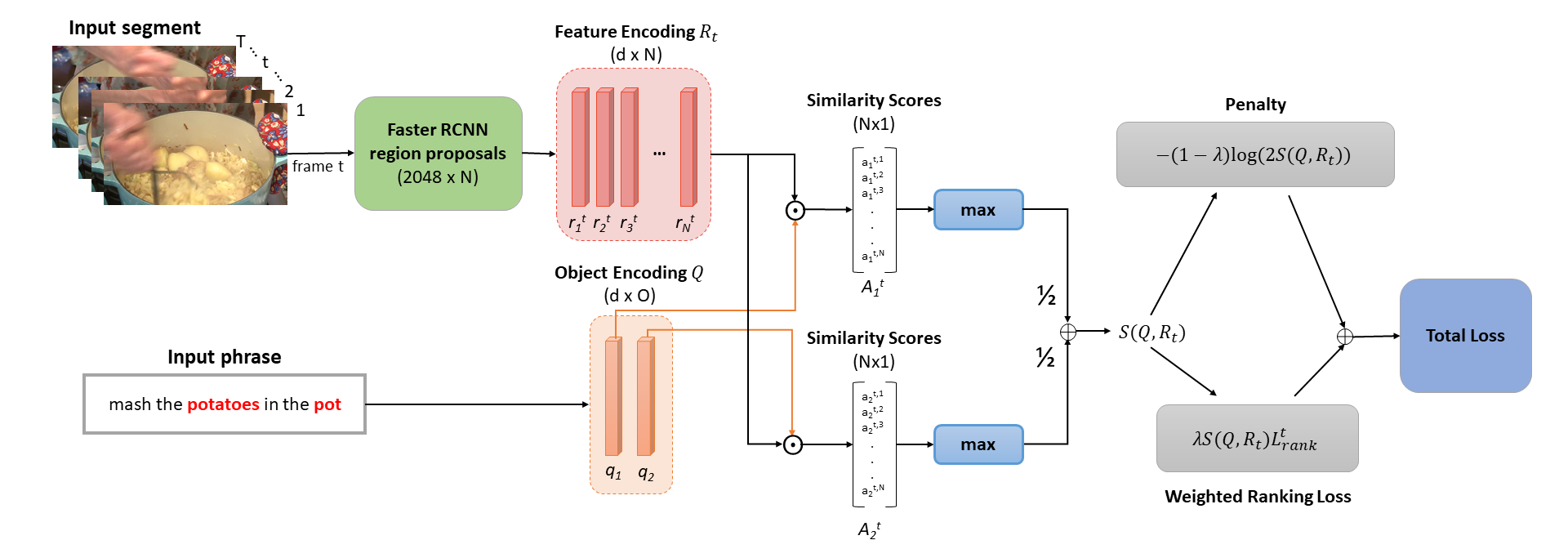}
\caption{An overview of our framework. Inputs to the system are a video segment and a phrase that describes the segment. The objects from the phrase are grounded for each sampled frame $t$. Object and proposal features are encoded to size $d$ and visual-semantic similarity scores are computed. 
The ranking loss is weighted by a confidence score which combined with the penalty form the final loss. The object relations are further encoded to guide the loss weights (see Sec.~\ref{sec:obj_interact} for details). During inference, the region proposal with the maximum similarity score with the object query is selected for grounding.}
\label{fig:model_overview}
\vspace{-10pt}
\end{figure}

\subsection{Frame-wise Loss Weighting}\label{sec:loss_weighting}
In our framework, each frame is considered separately to ground the same target objects. Fig. \ref{fig:model_overview} shows an overview of our model. We first estimate the likelihood that the query object is present in each video frame. If the likelihood is high, we judge the matching quality mainly on the matching loss (e.g., ranking loss). Otherwise, we down-weight the matching loss while bringing in a penalty loss. The lower the confidence, the higher the penalty. For clarity, we explain our idea when the matching loss is the ranking loss $L_{rank}$ but note that this can be generalized to other loss functions.

Let the ranking loss for frame $t$ be $L_{rank}^t$ and the similarity score between query $k$ and proposal $i$ be $a_k^{t,i}$. Let $Q=(q_1,\ldots,q_O)$ and $R_t=(r_1^t,\ldots,r_N^t)$.
We define the \textit{confidence score} of the prediction at frame $t$ as the visual-semantic matching score:
\begin{equation}
C_t = \frac{1}{O}\sum_{k=1}^O\max_{i}(a_k^{t,i}) \equiv S(Q, R_t),
\end{equation}
where $S(\cdot, \cdot)$ is defined in Eq.~\ref{eq:matching_score}. The corresponding \textit{penalty} is:
\begin{equation}
D_t = -\log(2C_t) = -\log[\frac{2}{O} \sum_{k=1}^O\max_{i}(a_k^{t,i})],
\end{equation}
inspired by~\cite{kendall2017multi}. The final loss for the segment is a weighted sum of frame-wise ranking losses and penalties:
\begin{small}
\begin{gather}\label{eq:final_loss}
L=\frac{1}{T}\sum_{t=1}^{T} [\lambda C_t L_{rank}^t + (1-\lambda)D_t], \\
L_{rank}^{t} = \sum_{R_t'\neq R_t}\sum_{Q'\neq Q} [\max(0,S(Q,R_t')-S(Q,R_t)+\Delta)+\max(0,S(Q',R_t)-S(Q,R_t)+\Delta)],
\end{gather}
\end{small}
where $\lambda$ is a static coefficient to balance the ranking loss and the penalty and can be validated on the validation set. A low $\lambda$ might cause the system to be over-confident on the prediction.

\subsection{Object Interaction}\label{sec:obj_interact}
We assume that the object types and their order in the language description can roughly determine when they appear in the video content, as motivated in Sec.~\ref{sec:intro}. We show that this language prior can work as the frame-wise confidence score. To consider the interaction among objects, we further encode each object query feature $q_k$ as:
\begin{gather}\label{eq:oi_loss}
J(q_k) = \text{MA}(q_k, Q, Q),
\end{gather}
where MA$(\cdot, \cdot, \cdot)$ is the multi-head self-attention layer~\cite{vaswani2017attention}, taking in the (query, key, value) triplet. It represents each query as the combination of all other queries based on their inter-relations. The built-in positional encoding layer~\cite{vaswani2017attention} in multi-head attention captures the order of objects appearing in the description. Note that the formulation is non-autoregressive, \ie, all the objects in the same description can interact with each other.

We evenly divide each video segment into $T'$ snippets and predict the confidence score for object $k$ to appear in each snippet based upon the concatenation of $J(q_k)$ and $q_k$. Note that $T'$ is a pre-specified constant that satisfies $T'\leq T$. The language-based confidence score $C_{lang}=(C_{lang}^1,\ldots,C_{lang}^{T'})$ is formulated as:
\begin{equation}\label{eq:cs}
C_{lang} = \frac{1}{O} \sum_{k=1}^O \text{Sigmoid}(W_{lang}[J(q_k);q_k]+b_{lang}),
\end{equation}
where $[\cdot \; ;\; \cdot]$ indicates the feature concatenation, $W_{lang}\in \mathbb{R}^{T'\times 2d}$ and $b_{lang}\in \mathbb{R}^{T'}$ are embedding weights and biases. We average the language-based and the similarity-based confidence score and rewrite Eq.~\ref{eq:final_loss} as:
\begin{equation}\label{eq:final_loss_interact}
L=\frac{1}{T}\sum_{t=1}^{T} [\lambda \frac{1}{2} (C_t+C_{lang}^{t_s}) L_{rank}^t - (1-\lambda)\log(C_t+C_{lang}^{t_s})]
\end{equation}
where $t_s=\min(\ceil{t/\ceil{\frac{T}{T'}}},T)$ is the snippet index and $\ceil{\cdot}$ stands for the ceiling operator.

\section{Experiments}\label{sec:experiments}
\subsection{Dataset} \label{sec:dataset}
\begin{figure}
\centering
\includegraphics[width=13cm]{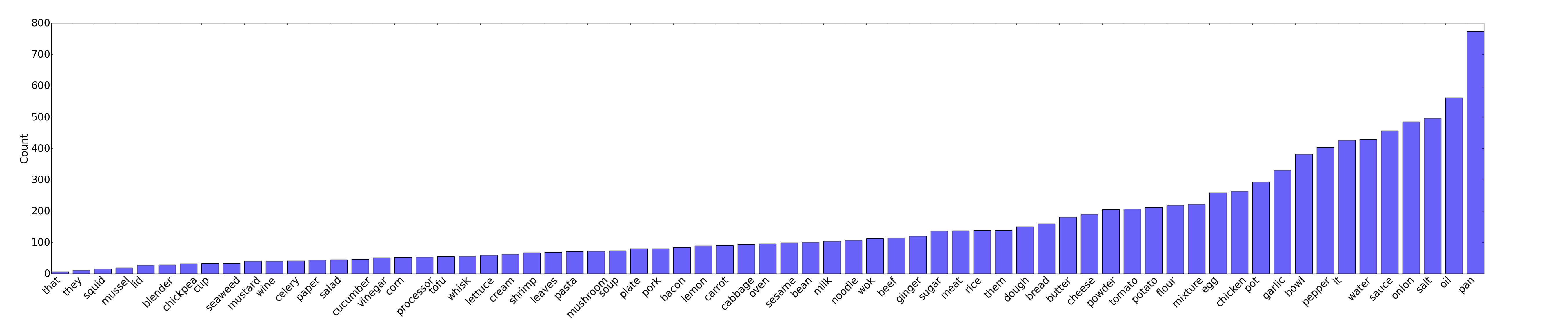}
\caption{Frequency count of each class label (including referring expressions).}
\label{fig:class_labels}
\vspace{-10pt}
\end{figure}

\noindent\textbf{YouCook2-BoundingBox.} YouCook2 \cite{zhou1703towards} consists of 2000 YouTube cooking videos from 89 recipes. Each video has recipe steps temporally annotated (\ie start timestamp and end timestamp) and each segment is described by a natural language sentence. The average segment duration is 19.6s. Our training set is the same as the YouCook2 training split, only paired sentences are provided. For each segment-description pair in the validation and testing set however, we provide bounding box annotations for the most frequently appearing objects from the dataset, \ie the top 63 recurring objects along with four referring expressions: \textit{it, them, that, they} (see Fig.~\ref{fig:class_labels}). These are used only during evaluation.

From YouCook2, we split each recipe step into a separate segment and sample it at 1 fps. We use Amazon Turk workers to draw bounding box around the objects in the video segment using the highlighted words in the sentence (from the 67 objects in our vocabulary). All annotations are further verified by the top 30 annotators.
Please see the Appendix for more details on annotations and quality control.

\vspace{-5pt}
\subsection{Baselines and Metrics}
\noindent\textbf{Baselines.}
We include two competitive baselines from published work: DVSA~\cite{karpathy2015deep} and GroundeR~\cite{rohrbach2016grounding}. DVSA is the Grounding by Ranking method which we build all our methods upon.
For fair comparison, all the approaches take in the same object proposals generated by Faster-RCNN~\cite{ren2017faster} (pre-trained on MSCOCO). Following the convention from ~\cite{karpathy2015deep,huang2018findingit}, we select the top $N=20$ proposals per frame and sample $T=5$ frames per segment unless otherwise specified. We also evaluate the Baseline Random, which chooses a random proposal as the output.

\noindent\textbf{Metrics.} We evaluate the grounding quality by bounding box localization accuracy (denoted as Box Accuracy). The output is positive if the proposed box has over 50\% IoU with the ground-truth annotation, otherwise negative. We compute accuracy for each object and average across all the object types.

\vspace{-5pt}
\subsection{Implementation Details}
The number of snippets $T'$ in Sec.~\ref{sec:obj_interact} is set to 5.
The encoding size $d$ is 128 for all the methods. Object labels are represented as one-hot vectors, which are encoded by a linear layer without the bias term. 
The loss factor $\lambda$ is cross-validated on the validation set and is set to 0.9. The ranking margin $\Delta$ is set to 0.1.
For training, we use stochastic gradient descent (SGD) with Nesterov momentum. The learning rate is set at 0.05 and the momentum is 0.9. We implement the model in PyTorch and train it using either a single Titan Xp GPU with SGD or 4 GPUs with synchronous SGD, depending on the validation accuracy. The model typically takes 30 epochs, \ie 4 hours to converge. More details are in the Appendix.

\vspace{-5pt}
\subsection{Results on Object Grounding}

The quantitative results on object grounding are shown in Tab.~\ref{tbl:accu-results}. The model with the highest score on the validation set is evaluated on the test split. 
We compute the upper bound as the accuracy when proposing all 20 proposals, to see how far the methods are from the performance limit. Note that the upper bound reported here is lower than that in~\cite{rohrbach2016grounding}. This is largely due to the domain shift from general scenes to cooking scenes and the large variance in our object states, e.g. zoom-in and zoom-out views, onions v.s. fried onion rings.

We show results on our proposed models, where the ``Loss Weighting'' model computes the confidence score with visual-semantic matching and the ``Object Interaction'' model computes the confidence score with textual guidance (Sec.~\ref{sec:obj_interact}).
Our full model averages these two scores as the final confidence score (Eq.~\ref{eq:final_loss_interact}).
The proposed methods demonstrate a steady improvement from the DVSA baseline, with a relative 1.40\% boost from loss weighting and another 1.62\% from combining object interaction, a total improvement of 3.02\%. On the other hand, the baseline has a higher validation score, which indicates model overfitting.
Note that text guidance alone (``Object Interaction'') works slightly worse than the baseline, showing that both visual and textual information are critical for inferring the frame-wise loss weights. Our methods also outperform other compared methods, GroundeR and Baseline Random by a large margin.

\setlength{\tabcolsep}{4pt}
\begin{table}[t]
\centering
\caption{Evaluation on localizing objects from the grounding-truth captions.}
\label{tbl:accu-results}
\begin{tabular}{lrr}
\hline
\noalign{\smallskip}
\multirow{2}{*}{Method} & \multicolumn{2}{r}{Box Accuracy (\%)} \\
& Val. & Test \\
\noalign{\smallskip}
\hline
\noalign{\smallskip}
\textbf{Compared methods} \\
Baseline Random &13.30 & 14.18\\
GroundeR~\cite{rohrbach2016grounding} & 19.63 & 19.94 \\
DVSA~\cite{karpathy2015deep} & 30.51 & 30.80\\
\hline
\textbf{Our methods} \\
Loss Weighting & 30.07 & 31.23\\
Object Interaction & 29.61 & 30.06 \\
Full Model & 30.31 & 31.73\\
\hline
\textbf{Upper bound} & 57.77 & 58.56\\
\hline
\end{tabular}
\end{table}
\setlength{\tabcolsep}{1.4pt}

\begin{figure}[t]
\centering
\includegraphics[width=0.45\textwidth]{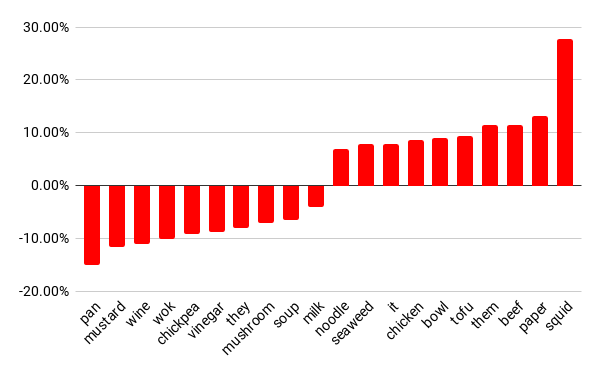}
\includegraphics[width=0.45\textwidth]{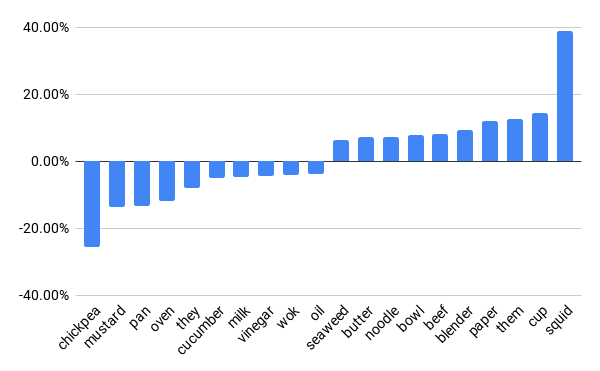}
\caption{Top 10 accuracy increases \& decreases by object category. (Left) Improvements of our Loss Weighting model over DVSA. (Right) Improvements of our Full Model over DVSA.}
\label{fig:improvements_by_cls}
\vspace{-10pt}
\end{figure}

\begin{figure}
\centering
\includegraphics[width=\textwidth]{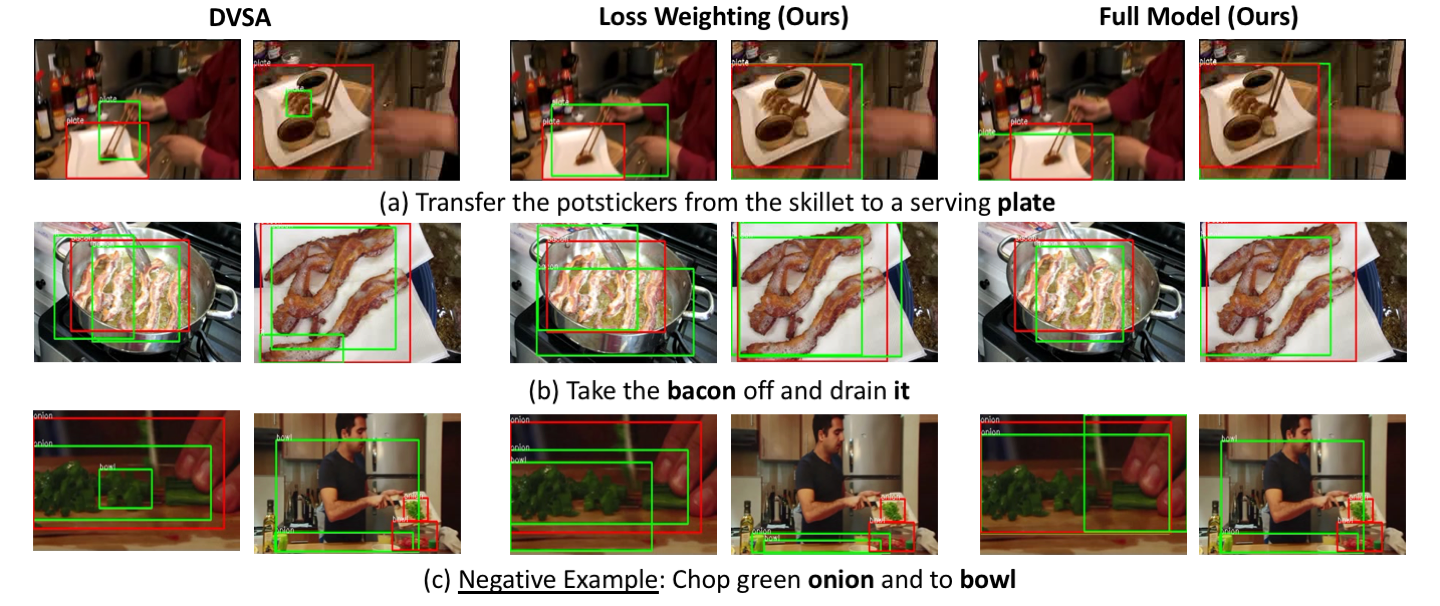}
\caption{Visualization of localization output from baseline DVSA and our proposed methods. Red boxes indicate ground-truths and green boxes indicate proposed regions. The first two rows show examples where our methods perform better than DVSA. The last row displays a negative example where all methods perform poorly. Better viewed in color.}
\label{fig:vis}
\vspace{-10pt}
\end{figure}

\noindent\textbf{Analysis.} We show in Fig.~\ref{fig:improvements_by_cls} the top 10 accuracy increases and decreases of our methods over the DVSA baseline, by object category. Our methods make better predictions on static objects such as ``squid'', ``beef'', and ``noodle'' and worse predictions on cookwares, such as ``wok'', ``pan'', and ``oven'', which involves more state changes, such as containing/not containing food or different camera perspectives. Our hypothesis is, our loss weighting framework favors consistent objects across frames, due to the shared frame-wise supervision. 

\noindent\textbf{Impact of Sampling Rate.} We investigate the impact of high video sampling rate on grounding accuracy by increasing the total number of frames per segment ($T$) from 5 to 20. The accuracy from DVSA drops from 30.80\% to 29.90\% and the accuracy from our Loss Weighted model drops from 31.23\% to 30.93\%.  We expected these inferior performances, due to the excessive object proposals. However, our loss weighted method only compromises 0.96\% of the accuracy while the accuracy from DVSA drops by 2.92\%, showing that our method is less sensitive to high sampling rate and predicts better on long frame sequences.

\noindent\textbf{Qualitative Results.} Fig.~\ref{fig:vis} visualizes the grounded objects with DVSA and our proposed methods. The first two rows show some positive examples. In Fig.~\ref{fig:vis} (a), we see with DVSA baseline the "plate" object is grounded to the incorrect regions in the frames. However our methods correctly select regions with a large IOU with the ground truth box. In Fig.~\ref{fig:vis} (b) the labels "bacon" and "it" refer to the same target object. Per our annotation requirements, there is only one ground truth box instead of two. The full model correctly combines both "bacon" and "it" grounds them to the same region proposal.
The last row that shows where all methods fail to ground the target objects adequately. This may be a result of errors in the top object proposals proposed since the scene is rather complicated. An additional explanation may be bias in the dataset, where during training the "bowl" object typically occupies the majority of the frame.

\noindent\textbf{Limitations.} There are two limitations in our method we hope to address in our future work. First, even though the frame-wise loss can to some degree enforce the temporal consistency between frames, we do not explicitly model the relation between frames, for instance motion information. The transition between object states across frames, e.g., raw meat to cooked meat, should be further studied. Second, our grounding performance is upper-bounded by the object proposal accuracy and we have no control over the errors from the proposals. An end-to-end version of the proposed method that solves both the proposing and the grounding problem can potentially improve the grounding accuracy.

\vspace{-5pt}
\section{Conclusion}
We propose a frame-wise loss weighted grounding model for video object grounding. Our model applies segment-level labels to the frames in each segment, while being robust to inconsistencies between the segment-level label and each individual frame.
We also leverage object interaction as textual guidance for grounding.
We evaluate the effectiveness of our models on the newly-collected video grounding dataset YouCook2-BoundingBox. Our proposed methods outperform competitive baseline methods by a large margin. Future directions include incorporating the video motion information and exploring an end-to-end solution for video object grounding.

\section*{Acknowledgement}
This work has been supported by DARPA FA8750-17-2-0112. This article solely reflects the opinions and conclusions of its authors but not DARPA. We thank Tianhang Gao, Ryan Szeto and Mohamed El Banani for their helpful discussions.

\newpage
\bibliography{egbib}

\begin{thebibliography}{30}
\providecommand{\natexlab}[1]{#1}
\providecommand{\url}[1]{\texttt{#1}}
\expandafter\ifx\csname urlstyle\endcsname\relax
  \providecommand{\doi}[1]{doi: #1}\else
  \providecommand{\doi}{doi: \begingroup \urlstyle{rm}\Url}\fi

\bibitem[Al-Omari et~al.(2017)Al-Omari, Duckworth, Hogg, and
  Cohn]{al2017natural}
Muhannad Al-Omari, Paul Duckworth, David~C Hogg, and Anthony~G Cohn.
\newblock Natural language acquisition and grounding for embodied robotic
  systems.
\newblock In \emph{AAAI}, pages 4349--4356, 2017.

\bibitem[Cinbis et~al.(2014)Cinbis, Verbeek, and Schmid]{cinbis2014multi}
Ramazan~Gokberk Cinbis, Jakob Verbeek, and Cordelia Schmid.
\newblock Multi-fold mil training for weakly supervised object localization.
\newblock In \emph{CVPR}, pages 2409--2416. IEEE, 2014.

\bibitem[Cirik et~al.(2018)Cirik, Berg-Kirkpatrick, and
  Morency]{cirik2018using}
Volkan Cirik, Taylor Berg-Kirkpatrick, and Louis-Philippe Morency.
\newblock Using syntax to ground referring expressions in natural images.
\newblock \emph{AAAI}, 2018.

\bibitem[Deselaers et~al.(2012)Deselaers, Alexe, and
  Ferrari]{deselaers2012weakly}
Thomas Deselaers, Bogdan Alexe, and Vittorio Ferrari.
\newblock Weakly supervised localization and learning with generic knowledge.
\newblock \emph{IJCV}, 100\penalty0 (3):\penalty0 275--293, 2012.

\bibitem[Divvala et~al.(2014)Divvala, Farhadi, and
  Guestrin]{divvala2014learning}
Santosh~K Divvala, Ali Farhadi, and Carlos Guestrin.
\newblock Learning everything about anything: Webly-supervised visual concept
  learning.
\newblock In \emph{CVPR}, pages 3270--3277, 2014.

\bibitem[Huang et~al.(2018)Huang, Buch, Dery, Garg, Fei-Fei, and
  Niebles]{huang2018findingit}
De-An Huang, Shyamal Buch, Lucio Dery, Animesh Garg, Li~Fei-Fei, and
  Juan~Carlos Niebles.
\newblock Finding ``it'': Weakly-supervised reference-aware visual grounding in
  instructional video.
\newblock \emph{To appear in CVPR}, 2018.

\bibitem[Karpathy and Fei-Fei(2015)]{karpathy2015deep}
Andrej Karpathy and Li~Fei-Fei.
\newblock Deep visual-semantic alignments for generating image descriptions.
\newblock In \emph{CVPR}, pages 3128--3137, 2015.

\bibitem[Karpathy et~al.(2014)Karpathy, Joulin, and Fei-Fei]{karpathy2014deep}
Andrej Karpathy, Armand Joulin, and Li~F Fei-Fei.
\newblock Deep fragment embeddings for bidirectional image sentence mapping.
\newblock In \emph{NIPS}, pages 1889--1897, 2014.

\bibitem[Kendall et~al.(2017)Kendall, Gal, and Cipolla]{kendall2017multi}
Alex Kendall, Yarin Gal, and Roberto Cipolla.
\newblock Multi-task learning using uncertainty to weigh losses for scene
  geometry and semantics.
\newblock \emph{arXiv preprint arXiv:1705.07115}, 2017.

\bibitem[Kwak et~al.(2015)Kwak, Cho, Laptev, Ponce, and
  Schmid]{kwak2015unsupervised}
Suha Kwak, Minsu Cho, Ivan Laptev, Jean Ponce, and Cordelia Schmid.
\newblock Unsupervised object discovery and tracking in video collections.
\newblock In \emph{ICCV}, pages 3173--3181. IEEE, 2015.

\bibitem[Lea et~al.(2016)Lea, Reiter, Vidal, and Hager]{lea2016segmental}
Colin Lea, Austin Reiter, Ren{\'e} Vidal, and Gregory~D Hager.
\newblock Segmental spatiotemporal cnns for fine-grained action segmentation.
\newblock In \emph{ECCV}, pages 36--52. Springer, 2016.

\bibitem[Ma et~al.(2017)Ma, Kadav, Melvin, Kira, AlRegib, and
  Graf]{ma2017attend}
Chih-Yao Ma, Asim Kadav, Iain Melvin, Zsolt Kira, Ghassan AlRegib, and
  Hans~Peter Graf.
\newblock Attend and interact: Higher-order object interactions for video
  understanding.
\newblock \emph{arXiv preprint arXiv:1711.06330}, 2017.

\bibitem[Ni et~al.(2016)Ni, Yang, and Gao]{ni2016progressively}
Bingbing Ni, Xiaokang Yang, and Shenghua Gao.
\newblock Progressively parsing interactional objects for fine grained action
  detection.
\newblock In \emph{CVPR}, pages 1020--1028, 2016.

\bibitem[Oquab et~al.(2015)Oquab, Bottou, Laptev, and Sivic]{oquab2015object}
Maxime Oquab, L{\'e}on Bottou, Ivan Laptev, and Josef Sivic.
\newblock Is object localization for free?-weakly-supervised learning with
  convolutional neural networks.
\newblock In \emph{CVPR}, pages 685--694, 2015.

\bibitem[Plummer et~al.(2015)Plummer, Wang, Cervantes, Caicedo, Hockenmaier,
  and Lazebnik]{plummer2015flickr30k}
Bryan~A Plummer, Liwei Wang, Chris~M Cervantes, Juan~C Caicedo, Julia
  Hockenmaier, and Svetlana Lazebnik.
\newblock Flickr30k entities: Collecting region-to-phrase correspondences for
  richer image-to-sentence models.
\newblock In \emph{ICCV}, pages 2641--2649. IEEE, 2015.

\bibitem[Plummer et~al.(2017)Plummer, Kordas, Kiapour, Zheng, Piramuthu, and
  Lazebnik]{plummer2017conditional}
Bryan~A Plummer, Paige Kordas, M~Hadi Kiapour, Shuai Zheng, Robinson Piramuthu,
  and Svetlana Lazebnik.
\newblock Conditional image-text embedding networks.
\newblock \emph{arXiv preprint arXiv:1711.08389}, 2017.

\bibitem[Prest et~al.(2012)Prest, Leistner, Civera, Schmid, and
  Ferrari]{prest2012learning}
Alessandro Prest, Christian Leistner, Javier Civera, Cordelia Schmid, and
  Vittorio Ferrari.
\newblock Learning object class detectors from weakly annotated video.
\newblock In \emph{CVPR}, pages 3282--3289. IEEE, 2012.

\bibitem[Ren et~al.(2017)Ren, He, Girshick, and Sun]{ren2017faster}
Shaoqing Ren, Kaiming He, Ross Girshick, and Jian Sun.
\newblock Faster r-cnn: towards real-time object detection with region proposal
  networks.
\newblock \emph{TPAMI}, 39\penalty0 (6):\penalty0 1137--1149, 2017.

\bibitem[Rohrbach et~al.(2016)Rohrbach, Rohrbach, Hu, Darrell, and
  Schiele]{rohrbach2016grounding}
Anna Rohrbach, Marcus Rohrbach, Ronghang Hu, Trevor Darrell, and Bernt Schiele.
\newblock Grounding of textual phrases in images by reconstruction.
\newblock In \emph{ECCV}, pages 817--834. Springer, 2016.

\bibitem[Rohrbach et~al.(2017)Rohrbach, Rohrbach, Tang, Oh, and
  Schiele]{rohrbach2017generating}
Anna Rohrbach, Marcus Rohrbach, Siyu Tang, Seong~Joon Oh, and Bernt Schiele.
\newblock Generating descriptions with grounded and co-referenced people.
\newblock \emph{arXiv preprint arXiv:1704.01518}, 3, 2017.

\bibitem[Song et~al.(2014)Song, Girshick, Jegelka, Mairal, Harchaoui, and
  Darrell]{song2014learning}
Hyun~Oh Song, Ross Girshick, Stefanie Jegelka, Julien Mairal, Zaid Harchaoui,
  and Trevor Darrell.
\newblock On learning to localize objects with minimal supervision.
\newblock \emph{arXiv preprint arXiv:1403.1024}, 2014.

\bibitem[Thomason et~al.(2017)Thomason, Padmakumar, Sinapov, Hart, Stone, and
  Mooney]{thomason2017opportunistic}
Jesse Thomason, Aishwarya Padmakumar, Jivko Sinapov, Justin Hart, Peter Stone,
  and Raymond~J Mooney.
\newblock Opportunistic active learning for grounding natural language
  descriptions.
\newblock In \emph{Conference on Robot Learning}, pages 67--76, 2017.

\bibitem[Vaswani et~al.(2017)Vaswani, Shazeer, Parmar, Uszkoreit, Jones, Gomez,
  Kaiser, and Polosukhin]{vaswani2017attention}
Ashish Vaswani, Noam Shazeer, Niki Parmar, Jakob Uszkoreit, Llion Jones,
  Aidan~N Gomez, {\L}ukasz Kaiser, and Illia Polosukhin.
\newblock Attention is all you need.
\newblock In \emph{NIPS}, pages 6000--6010, 2017.

\bibitem[Vondrick et~al.(2013)Vondrick, Patterson, and
  Ramanan]{vondrick2013efficiently}
Carl Vondrick, Donald Patterson, and Deva Ramanan.
\newblock Efficiently scaling up crowdsourced video annotation.
\newblock \emph{IJCV}, pages 184--204, 2013.

\bibitem[Xiao et~al.(2017)Xiao, Sigal, and Lee]{xiao2017weakly}
Fanyi Xiao, Leonid Sigal, and Yong~Jae Lee.
\newblock Weakly-supervised visual grounding of phrases with linguistic
  structures.
\newblock \emph{arXiv preprint arXiv:1705.01371}, 2017.

\bibitem[Yu and Siskind(2013)]{yu2013grounded}
Haonan Yu and Jeffrey~Mark Siskind.
\newblock Grounded language learning from video described with sentences.
\newblock In \emph{Proceedings of the 51st Annual Meeting of the Association
  for Computational Linguistics (Volume 1: Long Papers)}, volume~1, pages
  53--63, 2013.

\bibitem[Yu and Siskind(2017)]{yu2017sentence}
Haonan Yu and Jeffrey~Mark Siskind.
\newblock Sentence directed video object codiscovery.
\newblock \emph{International Journal of Computer Vision}, 124\penalty0
  (3):\penalty0 312--334, 2017.

\bibitem[Yu et~al.(2018)Yu, Lin, Shen, Yang, Lu, Bansal, and
  Berg]{yu2018mattnet}
Licheng Yu, Zhe Lin, Xiaohui Shen, Jimei Yang, Xin Lu, Mohit Bansal, and
  Tamara~L Berg.
\newblock Mattnet: Modular attention network for referring expression
  comprehension.
\newblock \emph{arXiv preprint arXiv:1801.08186}, 2018.

\bibitem[Zhou et~al.(2018{\natexlab{a}})Zhou, Xu, and Corso]{zhou1703towards}
Luowei Zhou, Chenliang Xu, and Jason~J Corso.
\newblock Towards automatic learning of procedures from web instructional
  videos.
\newblock \emph{AAAI}, 2018{\natexlab{a}}.

\bibitem[Zhou et~al.(2018{\natexlab{b}})Zhou, Zhou, Corso, Socher, and
  Xiong]{zhou2018end}
Luowei Zhou, Yingbo Zhou, Jason~J Corso, Richard Socher, and Caiming Xiong.
\newblock End-to-end dense video captioning with masked transformer.
\newblock \emph{arXiv preprint arXiv:1804.00819}, 2018{\natexlab{b}}.

\end{thebibliography}

\section*{Appendix}
\subsection*{More on Implementation Details}
When sampling frames from a segment, we evenly divide the segment into $T$ clips and randomly sample one frame from each clip as temporal data augmentation. The negative sample sentence $Q'$ is randomly sampled from all available sentences, but we exclude sentences that have overlapped objects with the positive sample $Q$. For self attention, we use a 2-layer 6-head multi-head attention module with the hidden size set to 256 and the dropout ratio set to 0.2.

For fair comparison, all the approaches take in the same object proposals generated by Faster-RCNN~\cite{ren2017faster}. The model is based upon ResNet-101 and pre-trained on MSCOCO for the object detection task.\footnote{ Details see \url{https://github.com/jwyang/faster-rcnn.pytorch}.} We take the 2048-dimension output after the RoI pooling as the region feature. We reduce the size of the region feature from 2048 to 128 with two linear layers, followed with dropout ($p=0.2$) and ReLU.

\subsection*{More on Data Annotation}
\noindent\textbf{Quality Control.} We use VATIC \cite{vondrick2013efficiently} as our annotation tool and Amazon Mechanical Turk (MTurk) as the crowdsourcing marketplace. To maintain quality control, a worker must annotate a gold-standard training video before being allowed to annotate the dataset. A gold-standard training video is an already annotated video segment that new workers are tested against. \cite{vondrick2013efficiently} introduced these videos to eliminate bad workers and limit annotation correction efforts. 
 A worker is not aware that they are completing a training video, but they are given unlimited attempts until it is successfully completed. All of the gold-standard training videos consists of three objects to be annotated and the worker must achieve an IoU of at least 50\% within every frame with one allowable mistake. The video segments were uploaded in batches, and with each new batch all workers were required to complete a different training video in order to continue annotating.
 We have a total of 94 annotators that completed the annotation tasks. The top 30 annotators (with the most accepted video segments) were selected to perform verification on the annotations. 

\noindent\textbf{Dataset Statistics} From the validation \& testing segments annotated we have a total of 4,325 annotated segments with 2,962 validation and 1,363 testing segments, respectively. These segments were extracted from 647 videos that contain words from our vocabulary list. 

\begin{figure}[t]
\centering
\includegraphics[width=0.32\textwidth]{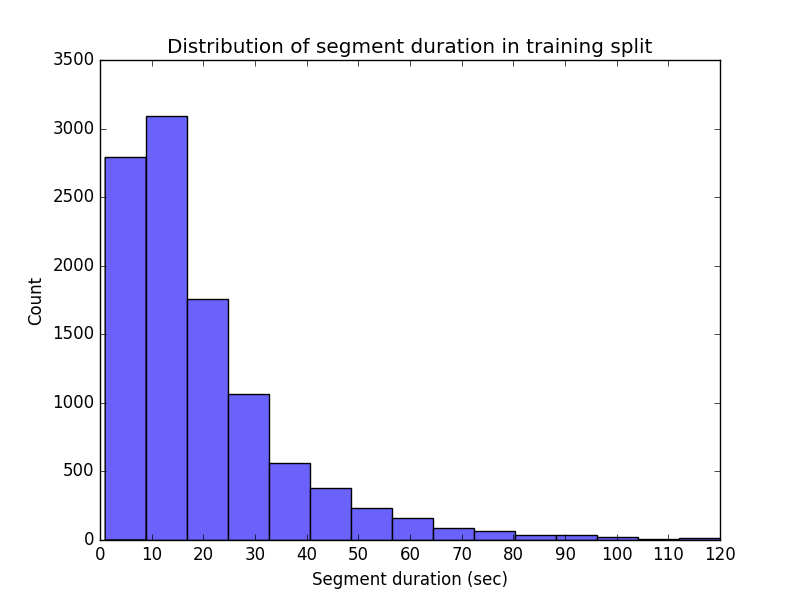}
\includegraphics[width=0.32\textwidth]{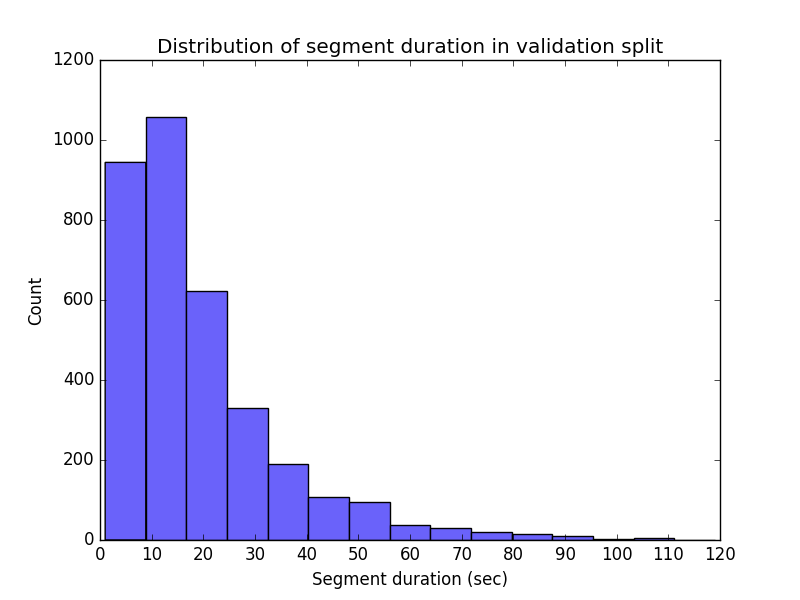}
\includegraphics[width=0.32\textwidth]{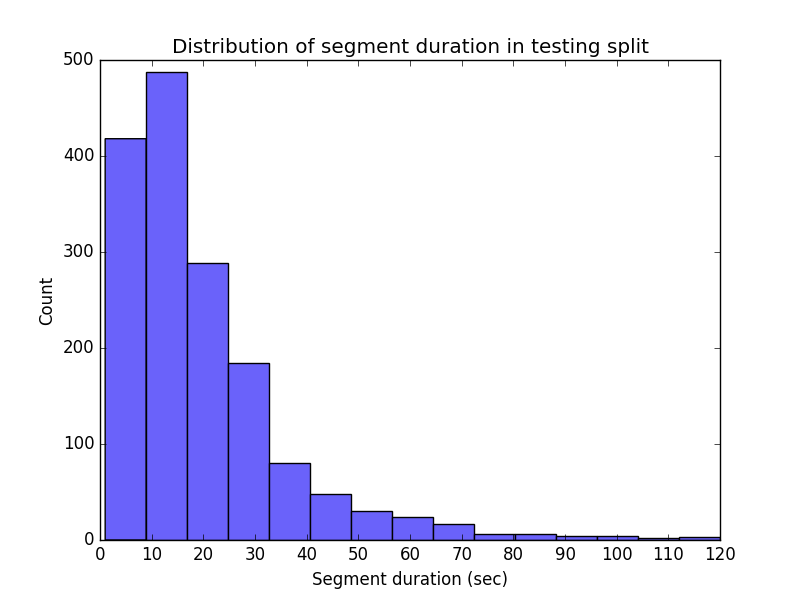}
\caption{Distribution of segment durations for train/val/test splits.}
\label{fig:duration_dist}
\end{figure}

\begin{figure}[t]
\centering
\includegraphics[width=0.32\textwidth]{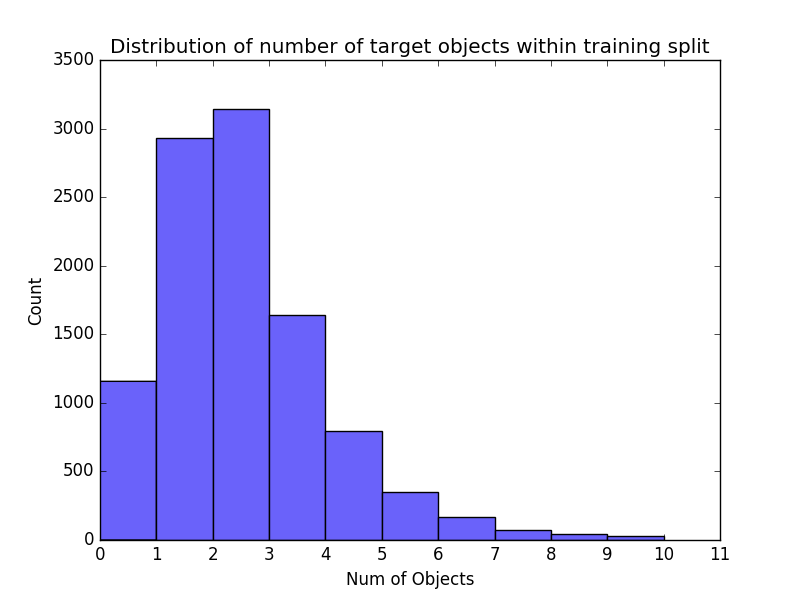}
\includegraphics[width=0.32\textwidth]{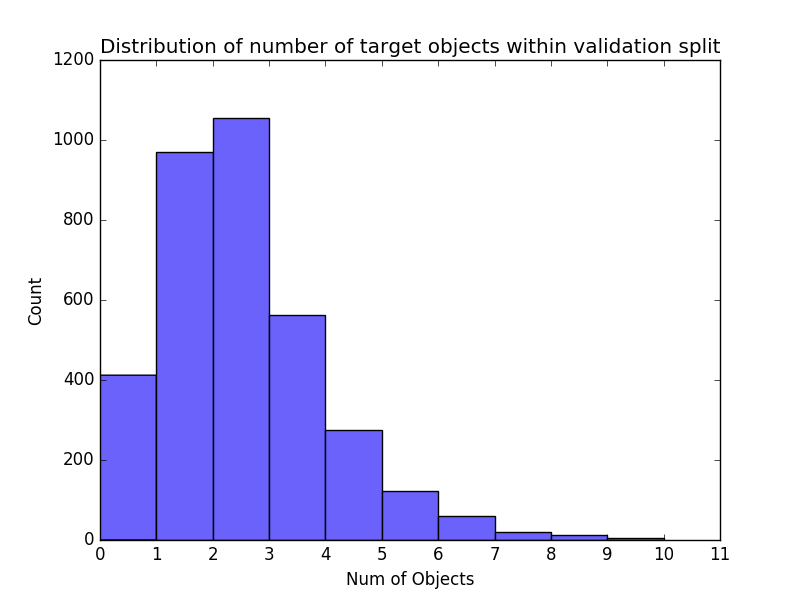}
\includegraphics[width=0.32\textwidth]{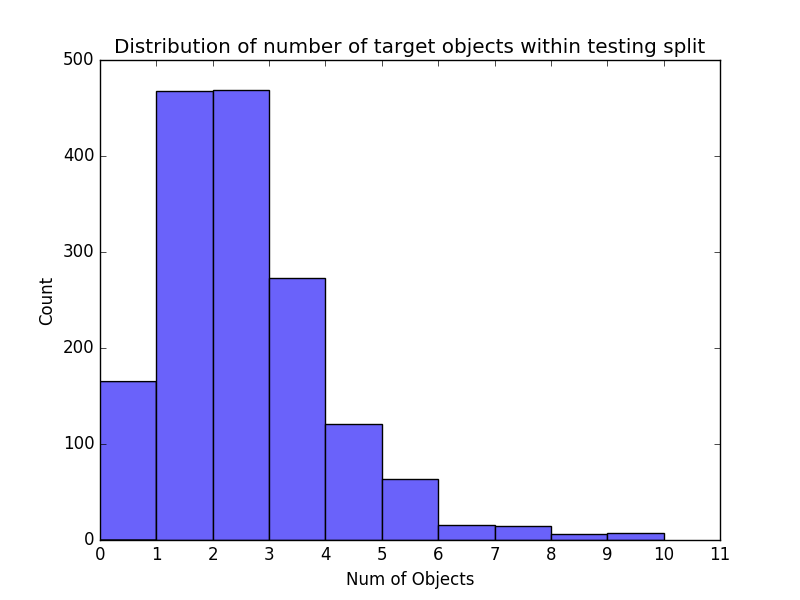}
\caption{Distribution of number of target objects within each segment for train/val/test splits. Target objects belong in our vocabulary of 67 words.}
\label{fig:target_dist}
\end{figure}

Fig. \ref{fig:duration_dist} shows the distribution of the segment durations from YouCook2, with mean and standard deviation of 19.6s and 18.2s across all splits.
Fig. \ref{fig:target_dist} displays the number of target objects from the annotated YouCook2-BoundingBox segments. The mean target object per sentence is 2.05 with a standard deviation of 1.49. The target objects are words that belong in our vocabulary list of 67 objects.

\begin{figure}[t]
\centering
\includegraphics[width=0.35\textwidth]{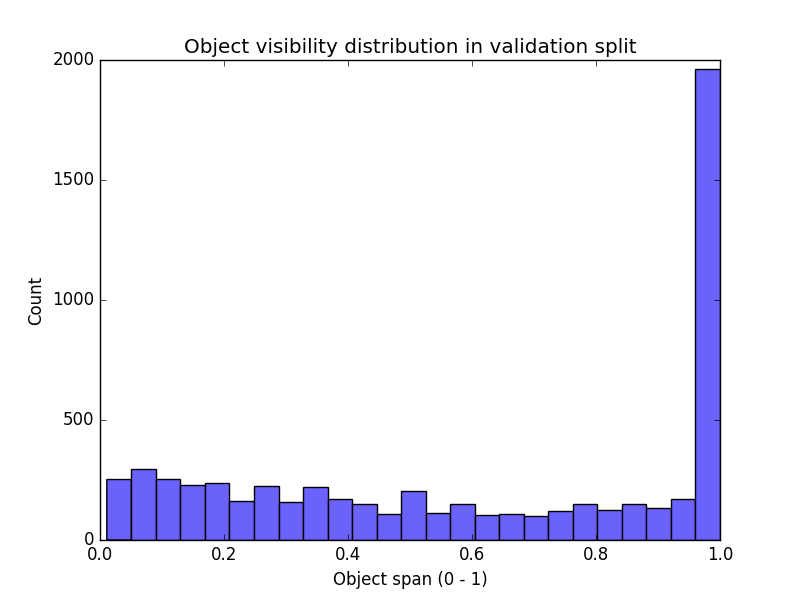}
\includegraphics[width=0.35\textwidth]{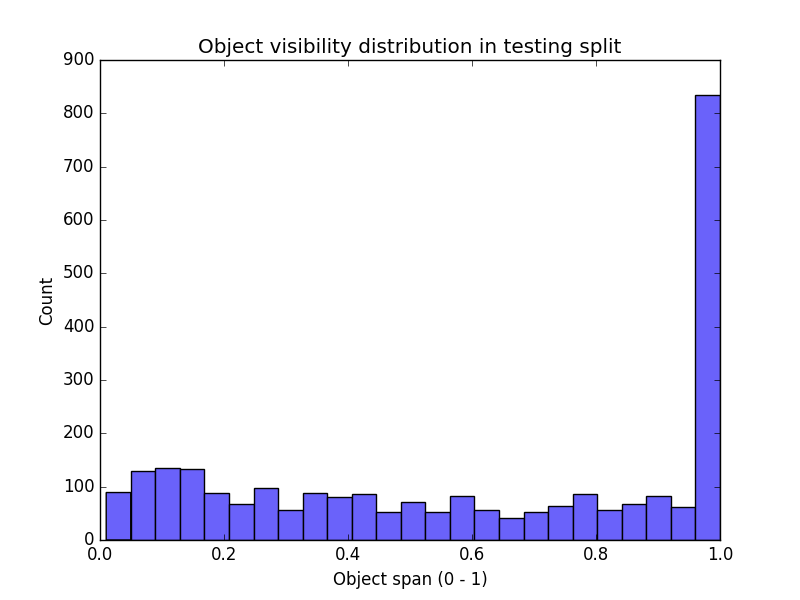}
\caption{Span of object duration in each segment for annotated val/test splits.}
\label{fig:visibility_dist}
\end{figure}

When completing the annotations, the workers were given the option to mark an object as "outside of view frame", "occluded", or both. We define an object's visibility as in view of the current frame with no occlusion. From our collected annotations, Fig. \ref{fig:visibility_dist} shows each object's visibility duration in the validation \& testing split. In the validation split objects are visible 60.72\% of the time, and 60.58\% for testing. Note from Fig. \ref{fig:visibility_dist} there is a spike in objects with 100\% duration, this is attributed to the shorter segments from our collected data. It is perfectly reasonable to have a visible object for the entire duration of shorter segments, some as short as 2 seconds.

\begin{figure}[t]
\includegraphics[width=0.9\textwidth]{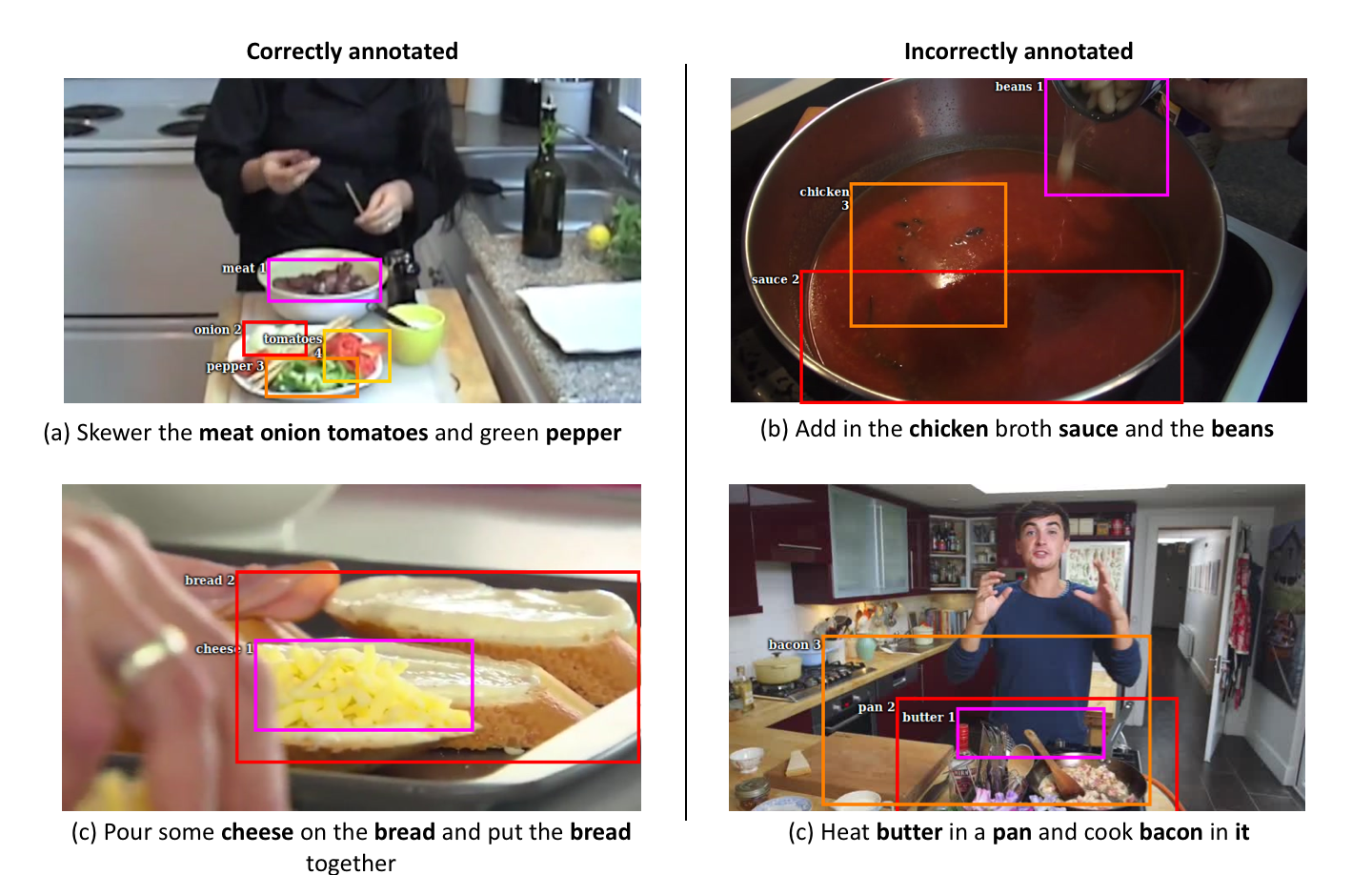}\centering
\caption{Annotations completed by MTurk workers; The images on the left denote correct annotations and the right shows incorrect annotations. Each image is a frame from the video segment accompanied with its descriptive phrase. Better viewed in color.}
\label{fig:ann_example}
\end{figure}

\clearpage
\section*{Errata}
After releasing the original version of the results, we discovered an error in the calculation of the evaluation metric (i.e., a scaling issue in the object proposal coordinates). This later version fixes that error. For completeness, we include the tables from both cases here for comparison (Tab. 1 for the initial results and Tab. 2 for the updated results). We note the performance ordering does not change, that all methods see a significant rise with respect to the baseline and the relative performance improvement decreases.

\setlength{\tabcolsep}{4pt}
\begin{table}[h]
\centering
\caption{(\textbf{Initial.}) Evaluation on localizing objects from the grounding-truth captions.}
\label{tbl:errata1}
\begin{tabular}{lrr}
\hline
\noalign{\smallskip}
\multirow{2}{*}{Method} & \multicolumn{2}{r}{Box Accuracy (\%)}\\
& Val. & Test \\
\noalign{\smallskip}
\hline
\noalign{\smallskip}
\textbf{Compared methods} \\
Baseline Random & 8.60 & 9.51 \\
GroundeR~\cite{rohrbach2016grounding} & 12.91 & 13.72 \\
DVSA~\cite{karpathy2015deep} & 14.70 & 16.85 \\
\hline
\textbf{Our methods} \\
Loss Weighting & 15.80 & 17.74 \\
Object Interaction & 14.86 & 16.33 \\
Full Model & 15.83 & 18.39 \\
\hline
\textbf{Upper bound} & 45.97 & 47.17 \\
\hline
\end{tabular}
\end{table}
\setlength{\tabcolsep}{1.4pt}

\setlength{\tabcolsep}{4pt}
\begin{table}[h]
\centering
\caption{(\textbf{Updated.}) Evaluation on localizing objects from the grounding-truth captions.}
\label{tbl:errata2}
\begin{tabular}{lrr}
\hline
\noalign{\smallskip}
\multirow{2}{*}{Method} & \multicolumn{2}{r}{Box Accuracy (\%)} \\
& Val. & Test \\
\noalign{\smallskip}
\hline
\noalign{\smallskip}
\textbf{Compared methods} \\
Baseline Random &13.30 & 14.18\\
GroundeR~\cite{rohrbach2016grounding} & 19.63 & 19.94 \\
DVSA~\cite{karpathy2015deep} & 30.51 & 30.80\\
\hline
\textbf{Our methods} \\
Loss Weighting & 30.07 & 31.23\\
Object Interaction & 29.61 & 30.06 \\
Full Model & 30.31 & 31.73\\
\hline
\textbf{Upper bound} & 57.77 & 58.56\\
\hline
\end{tabular}
\end{table}
\setlength{\tabcolsep}{1.4pt}

\end{document}